\documentclass{ieeeaccess}
\usepackage{cite}
\usepackage{amsmath,amssymb,amsfonts}
\usepackage{algorithmic}
\usepackage{graphicx}
\usepackage{textcomp}
\usepackage{tabularx}
\usepackage{arydshln}
\usepackage{caption}

\usepackage{lipsum}
\usepackage{multirow}
\usepackage{amsmath,amssymb}
\usepackage{color}
\usepackage{kotex}
\usepackage[normalem]{ulem}

\usepackage{pifont}
\newcommand{\cmark}{\ding{51}}%

\usepackage{makecell}
\usepackage[symbol]{footmisc}
\usepackage{hyperref}

\hypersetup{
    colorlinks=true,
    linkcolor=blue,
    filecolor=magenta,      
    urlcolor=magenta,
    pdftitle={Overleaf Example},
    pdfpagemode=FullScreen,
    }

\DeclareCaptionFont{ieeeblue}{\color{accessblue}}
\DeclareCaptionLabelFormat{myformat}{\figcapfont{\textbf{#1}\textbf{#2}}}
\def\BibTeX{{\rm B\kern-.05em{\sc i\kern-.025em b}\kern-.08em
    T\kern-.1667em\lower.7ex\hbox{E}\kern-.125emX}}
\begin{document}
\history{Date of publication xxxx 00, 0000, date of current version xxxx 00, 0000.}
\doi{10.1109/ACCESS.2017.DOI}

\title{Self-Supervised Visual Learning \\by Variable Playback Speeds Prediction of a Video}
\author{\uppercase{Hyeon Cho}\authorrefmark{1}, \uppercase{Taehoon Kim}\authorrefmark{1},
\uppercase{Hyungjin Chang}\authorrefmark{2}, 
\uppercase{and Wonjun Hwang}.\authorrefmark{1}, \IEEEmembership{Member, IEEE}}
\address[1]{Department of Artificial Intelligence, Ajou University, 206, World cup-ro, Yeongtong-gu, Suwon-si, Gyeonggi-do, Republic of Korea (e-mail: \{ch0104, th951115, wjhwang\}@ajou.ac.kr)}
\address[2]{The School of Computer Science, University of Birmingham, Edgbaston, Birmingham, B15 2TT, United Kingdom (e-mail: h.j.chang@bham.ac.uk)}
\tfootnote{“This research was partially supported by the NRF(National Research Foundation) program(NRF-2020R1F1A1066049) and the ITRC(Information Technology Research Center) support program(IITP-2021-2018-0-01431)”}

\markboth
{H.Cho \headeretal: Self-Supervised Visual Learning by Variable Playback Speeds Prediction of a Video}
{H.Cho \headeretal: Self-Supervised Visual Learning by Variable Playback Speeds Prediction of a Video}

\corresp{Corresponding author: Wonjun Hwang (e-mail: wjwhang@ajou.ac.kr).}

\begin{abstract}
We propose \textit{a self-supervised visual learning method} by predicting the variable playback speeds of a video. Without semantic labels, we learn the spatio-temporal visual representation of the video by leveraging the variations in the visual appearance according to different playback speeds under the assumption of temporal coherence. To learn the spatio-temporal visual variations in the entire video, we have not only predicted a single playback speed but also generated clips of various playback speeds and directions with randomized starting points. Hence the visual representation can be successfully learned from the meta information (playback speeds and directions) of the video. We also propose a new layer-dependable temporal group normalization method that can be applied to 3D convolutional networks to improve the representation learning performance where we divide the temporal features into several groups and normalize each one using the different corresponding parameters. We validate the effectiveness of our method by fine-tuning it to the action recognition and video retrieval tasks on UCF-101 and HMDB-51.
\footnote{All the source code is released in \href{https://github.com/hyeon-jo/PSPNet}{ https://github.com/hyeon-jo/PSPNet}.}
\end{abstract}


\titlepgskip=-15pt

\maketitle

\section{Introduction}

\noindent The outstanding performance of image-based applications such as image recognition~\cite{ResNet}, object detection~\cite{Yolo}, and image segmentation~\cite{SegNet} rely on large amounts of annotated data; for example, ImageNet~\cite{ImageNet}, is used to train the deep-stacked layers of a convolutional neural network (CNN). However, when studying video recognition using deep learning, the availability of large sets of annotated data, such as Kinetics~\cite{Kinetics}, is extremely costly and laborious. Therefore, an increasing need has arisen for a method that can be adapted to new domains without leveraging a huge amount of expensive supervision. 

\begin{figure}[t]
\centering
\includegraphics[height=4.7cm]{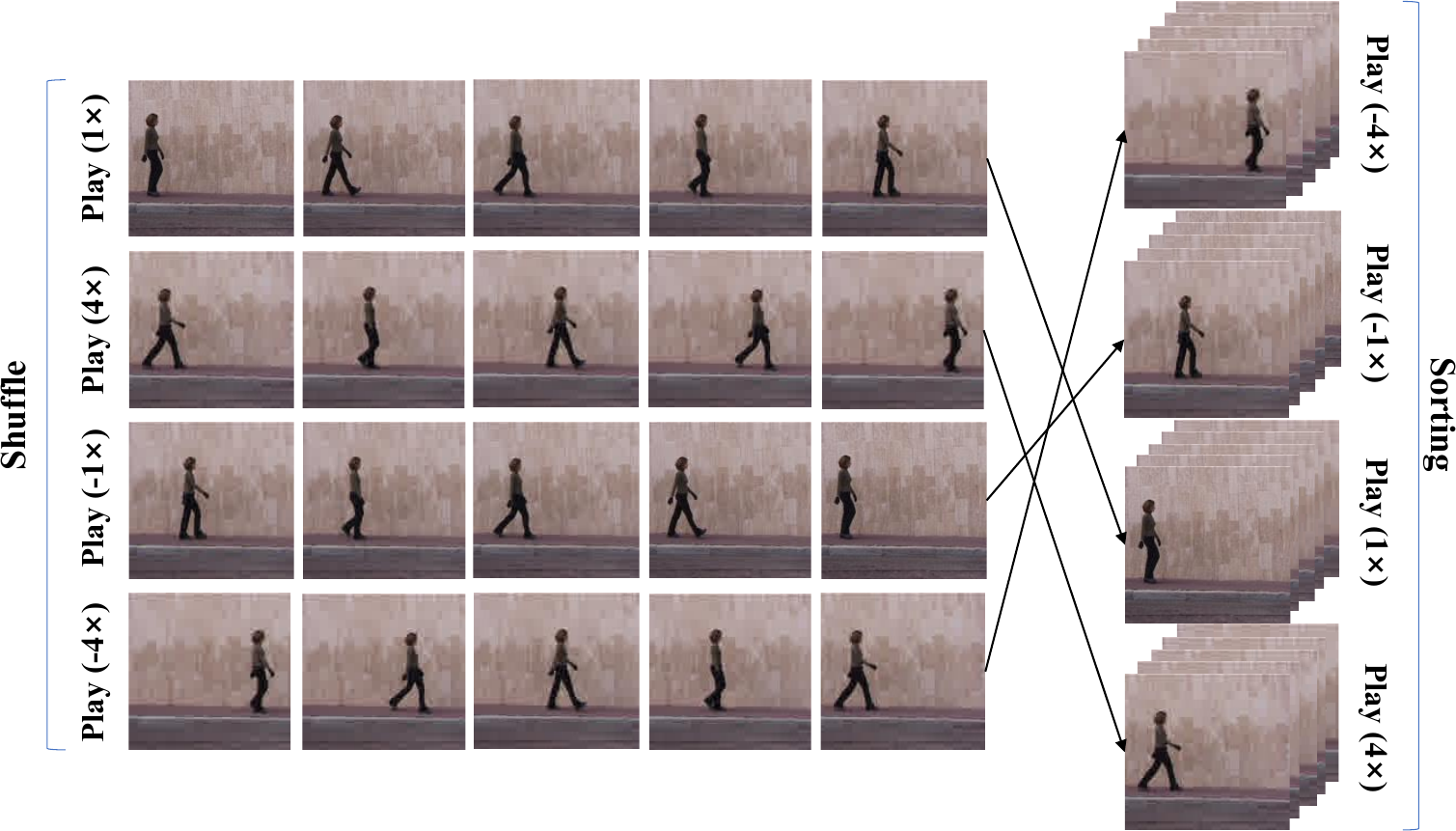}
\caption{\textbf{Conceptual illustration of the proposed method.} We play back the video at variable speeds such as -4x, -1x, 1x and 4x with randomized starting points to create three clips. After shuffling the clips, we predict the playback speed of each clip and sort them according to the correct playback speeds. Once 3D ConvNets are able to solve this surrogate task without additional information, the network is ready to understand the temporal dynamics of the videos.}
\label{fig:01}
\vspace{-5mm}
\end{figure}

Recently, self-supervised learning has attracted increasing attention in many classification tasks such as image classification using a jigsaw puzzle~\cite{Noroozi2016}, predicting the rotation of images~\cite{Gidaris2018}, video classification based on prediction of the frame order~\cite{HLee2017} and clip order~\cite{DXu2019}, and domain adaptation using self-supervised learning instead of the gradient reversal layer~\cite{SS_DA}. The self-supervised learning is a common learning framework that makes use of surrogate tasks that can be formulated without supervision by corresponding labels. The data itself could provide the supervisory signal for learning the image representation via self-supervised learning. 

From the viewpoint of video classification, video data has its inherent characteristics such as spatial and temporal coherence, and most of the video-based self-supervised learning methods leverage spatio-temporal representation learning based on the steady chronological order of video data to understand the underlying temporal dynamics~\cite{HLee2017, DXu2019,DLou2020}. Specifically, the frames selected at equal temporal intervals are shuffled and their chronological order predicted~\cite{HLee2017}, whereas clips instead of video frames are used to leverage the inner steady temporal dynamics~\cite{DXu2019}. However, these previous methods have learned the visual representation of video via the temporal dynamics, but they make use limited data randomly selected from the entire video. Although this approach is advantageous in that it strengthens learning efficiency without supervision, the reliance on limited data is a constraint that has hampered further improvements in video classification performance.

In this paper, we propose a new RGB-based pretext task of self-supervised learning that predicts the order of various speeds and directions of video. In addition, we also suggest layer-dependable temporal group normalization (TGN) that helps 3D ConvNets learn better temporal dynamics by replacing vanilla batch normalization (BN). First, the video data are played repeatedly from the random start points at various predefined speeds including fast forward playback or backward playback. Each time the selected frames are collected to create a fixed-length clip. By recognizing the relative speeds of the clips, the variations in appearance and temporal coherence can be learned efficiently in the video over time without the semantic labels. Specifically, as shown in Fig~\ref{fig:01}, rather than simply predicting the chronological order of the frames, we formulate a task that predicts multiple playback speeds and then sorts the clips according to the playback speed. This approach enables the proposed method to learn the temporal video dynamics using as much frame data as possible from the video. 

The main contributions of this paper can be summarized as follows: 
\begin{itemize}
    \item We introduce a self-supervised spatio-temporal representation learning via predicting the order of the multiple forward and backward playback speeds, which plays a pivotal role in generating many types of clips from videos and learning the spatio-temporal structure of videos without any manual annotations.
    \item We propose a new layer-dependable temporal group normalization method which enables efficient 3D ConvNet learning under the large variations of appearance and temporal coherence at videos. 
    \item We show that using the proposed method improves the self-supervised learning performance for action recognition and video retrieval tasks through the extensive experiments on various 3D ConvNet architectures and datasets.
\end{itemize}

\section{Related Work}
In this section, we review relevant literature on self-supervised learning and batch normalization. Self-supervised learning representation~\cite{Doersch2015, Noroozi2016, RZhang2016, XWang2017, Pathak2016, Gidaris2018} has been studied for leveraging large-scaled training data without the label information in many vision applications such as image classification~\cite{Noroozi2016}, Generative Adversarial Network~\cite{TChen2018} or Domain Adaptation~\cite{YSun2019} and video classification~\cite{DXu2019}. For this purpose, the surrogate tasks are formulated such that an inherent visual representation needs to be learned without a supervisory signal, and it plays a pivotal role in the accuracy of the vision applications.

\begin{figure*}[t]
\centering
\includegraphics[height=8.0cm]{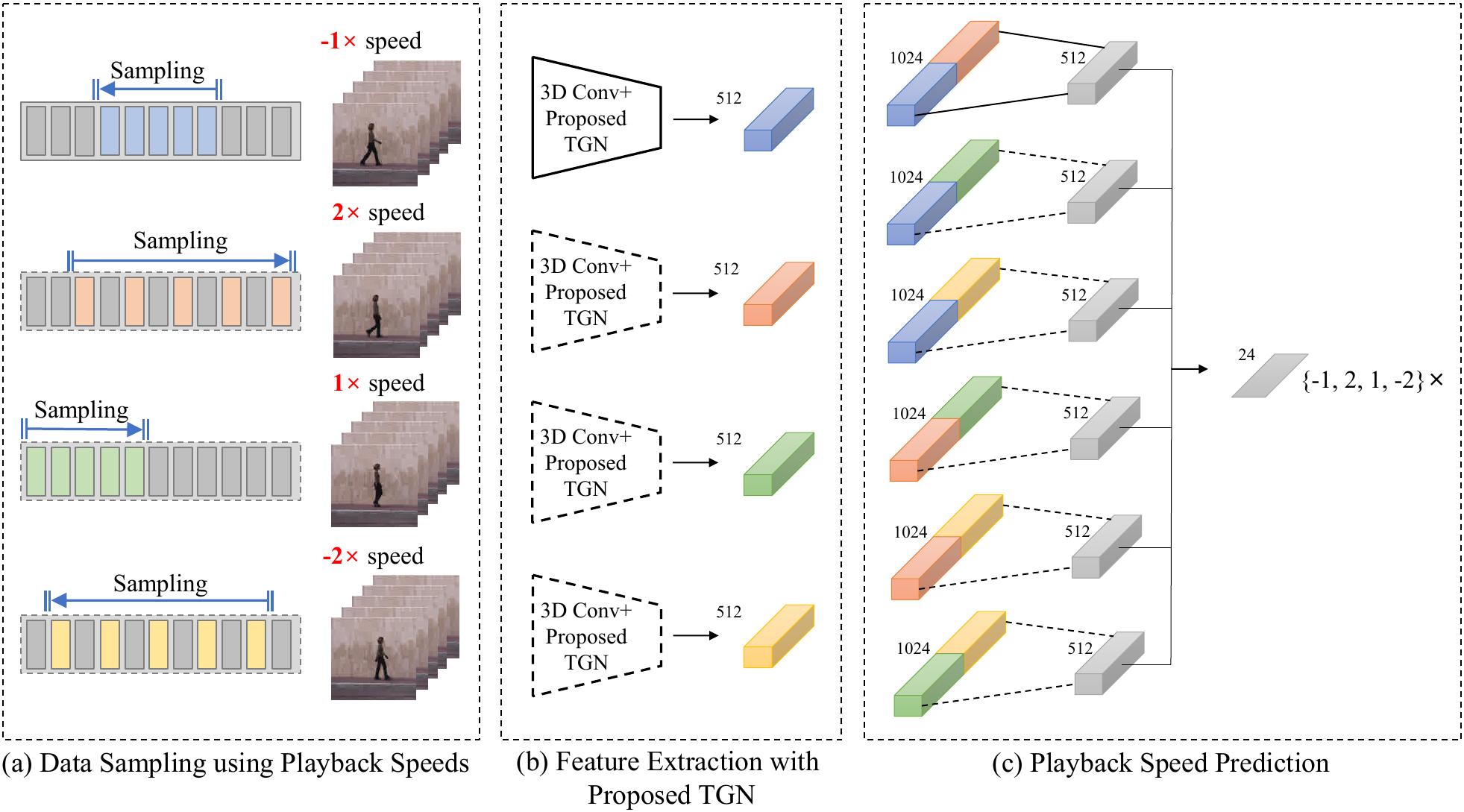}
\caption{\textbf{The overall framework of the proposed method}. (a) Data sampling using playback speeds. We sample frames according to the different playback speeds from the random initial points to form a clip and randomly shuffle them for 3D ConvNet. (b) Feature extraction with layer-dependable TGN. For efficient video learning, we train a 3D ConvNet using the proposed TGN instead of batch normalization. (c) Playback speed prediction. The extracted features are concatenated pairwise. The fully connected layers are used to encode the features, and the final layer uses these features to predict the playback speeds of the input clips. The dashed lines indicate that the network weights are shared with the straight lines.}
\label{fig:02}
\vspace{-5mm}
\end{figure*}

\textbf{Self-supervised learning for static images: }
Self-supervised learning representation~\cite{Doersch2015, Noroozi2016, RZhang2016, XWang2017, Pathak2016, Gidaris2018} has been studied for leveraging large-scaled training data without the label information in many vision applications such as image classification~\cite{Noroozi2016}, Generative Adversarial Network~\cite{TChen2018} or Domain Adaptation~\cite{YSun2019} and video classification~\cite{DXu2019}. For this purpose, the surrogate tasks are formulated such that an inherent image representation needs to be learned without a supervisory signal, and it plays a pivotal role in the accuracy of the vision applications. Doersch et al.~\cite{Doersch2015} designed a method to learn the representation of images by predicting the relative locations of two randomly sampled image patches. The rule of the jigsaw puzzle~\cite{Noroozi2016} has been generally employed for predicting a permutation of multiple randomly sampled patches. Another approach is identifying the randomly shuffled channels of the color image for image colorization~\cite{RZhang2016}. The cue of the transitive in variance~\cite{XWang2017} could be used to match the patches of an image as for another self-supervised learning trial. Recently, Gidaris et al.~\cite{Gidaris2018} proposed a straightforward self-supervised learning method for predicting the angle of the random rotation transformation of an image and achieved good results in many ways.

\textbf{Self-supervised learning for videos: }
Looking at the video-based self-supervised learning-based methods, temporal order verification~\cite{IMisra2016} was early proposed for leveraging the temporal order of the sequential images because the video frames are stored in chronological order. It simply checked whether the temporal order is correct or not without the semantic labels. Soon after, Lee et al.~\cite{HLee2017} proposed the frame order sorting method, which increased the performance of the self-supervised learning method in video recognition. They formulated the sequence sorting task as revealing the underlying chronological order of the sampled frames. In~\cite{Fernando2017}, the odd-one-out network has been proposed for self-supervised video representation learning, which identifies the odd temporal order of frames among some trials. Wang et al.~\cite{JWang2019} proposed the self-supervised learning method that learns visual features by regressing both motion and appearance statics along the spatial and temporal dimension. The jigsaw puzzle~\cite{Noroozi2016} was extended to the space-time cubic puzzles for training a 3D ConvNet in~\cite{DKim2019}. Xu et al.~\cite{DXu2019} have efficiently improved the frame order prediction method~\cite{HLee2017} by sorting the order of the neighboring clips as known as video clip order prediction (VCOP), where the clips are consistent with the video dynamics. In~\cite{DLou2020}, the video cloze procedure (VCP) was proposed to learn the spatial-temporal representation of video data based on a method that uses spatial rotations and temporal shuffling method, which enhanced the accuracy in action recognition. Our proposed method is inspired by~\cite{HLee2017} and~\cite{DXu2019}, but we make use of the playback speeds of the videos, not the correct sequential order of sampled frames~\cite{HLee2017} or clips~\cite{DXu2019}. This approach provides rich self-supervision to improve the learning performance based on the use of video.
Inspired by SlowFast Network~\cite{SlowFast}, the latest studies~\cite{oops2020,speednet2020,temporaltransforms2020} on video speed have been conducted. 
\cite{oops2020} proposed a pretext method to predict the normal video speed to detect an unintentional event. 
In ~\cite{speednet2020}, they introduced a pretext for predicting the binary speediness (e.g., fast or normal speed) of moving objects in videos. 
Jenni et al~\cite{temporaltransforms2020} showed a pretext learning not only the video speed but also temporal transformations.
On the other hand, we design a pretext based on the fundamental temporal features of video playback speeds order prediction, e.g., playback speeds and directions.

\textbf{Normalization methods: }
Batch normalization has shown considerable progress from the viewpoint of efficient learning of deep learning. After the success of batch normalization~\cite{BN,BN2}, where the mean and variance are used for global normalization along the batch dimension, many normalization methods~\cite{LN,IN,WN,GN} have been proposed to improve the performance. First, weight normalization~\cite{WN} is suggested normalizing the filter weights, and layer normalization~\cite{LN} performed the normalization along the channel dimension only. Instance normalization~\cite{IN} operates along with each instance sample. Group normalization~\cite{GN} (GN) is a compromise between layer normalization and instance normalization where they proposed a layer that divides channels into groups and normalizes the features within each group. However, all these studies have limited performance in terms of the normalization along with channel, layer, or instance features, without deep consideration of the temporal features in the video. In this respect, we extend the group normalization to the layer-dependable TGN for video recognition with a novel task of self-supervised learning. 

\section{Our Approach}
We propose a surrogate task using variable playback speed prediction and 3D ConvNet using layer-dependable TGN to enable a large number of unlabeled videos to be used for efficient learning. When a 3D ConvNet is used to solve the Playback Speed Prediction (PSP) task, the 3D ConvNet successfully learns the fundamental visual representation of videos by understanding the temporal coherence changes according to the different playback speeds. For this purpose, as summarized in Fig~\ref{fig:02}, the proposed method consists of data sampling using playback speeds, the feature extraction with the proposed layer-dependable TGN, and the PSP network.

\subsection{Data sampling using variable playback speeds}
From a single video, we sample frames based on the the different playback speeds (e.g., from -5$\times$ to 5$\times$) and generate clips of the various playback speeds with the sampled frames. Subsequently, the clips are randomly shuffled as inputs of the 3D ConvNet. Note that the starting frame is always set to an arbitrary position during sampling so that more diverse frames can be selected from one video. As described in \cite{SlowFast}, the multiple clips with different playback speeds could potentially allow the spatial semantics to be learned at slow speed and the motion dynamics to be learned at fast speed. Besides, because the clips are made by playing forward and backward at different speeds, it is possible to learn the temporal dynamics successfully even if the directional movements (e.g., push or pull) of the target in the video are fast or slow. 


We define a tuple of $n$ clips as $L_n$ as follows:
\begin{equation}
    L_n=
    \begin{cases} 
    \{l_{+3},l_{-3}\} & \text{if $n$ = 2}\\
    \bigcup_{i=1}^{(n-1)/2}\{l_{+1}, l_{+(2i+1)}, l_{-(2i+1)}\} & \text{if $n$ is odd}\\
    L_{n-1}\cup\{l_{-1}\} & \text{if $n$ is even}\\
    \end{cases}
    \label{eq:2}
\end{equation}
where $n$ is the number of clips, and the subscript $s$ of the clip $l_s$ is the playback speed or the frame rate. Positive playback speed indicates fast forward playing, and negative speed is reverse playing or rewinding. When constructing a tuple of clips, we always ensure that half of the clips have positive speed and the other half have negative speed. Therefore, our formulation can accurately represent the direction of dynamics. In particular, our proposing method can distinguish between open and close dynamics which has been difficult to achieve so far~\cite{HLee2017}. 
In the order prediction, there are $n!$ ($n$ factorial)
possibilities in total. The clip, $l$, consists of $m$ frames from random initial frame, $f_i$, (where $f_i$ is the $i^{th}$ frame of the original video), which is derived by
\begin{align}
    l_{+s}&=\{f_i,f_{i+s},f_{i+2s},...,,f_{i+(m-1)s}\},\\
    l_{-s}&=\{f_{i-(m-1)s},f_{i-(m-2)s},...,,f_{i}\}.
    \label{eq:34}
\end{align}

Unlike \cite{DXu2019} we allow the same frames to be selected between the clips because of random initialization and the different frame rate associated with the clips. This allowance gives us a higher degree of freedom in using videos, so that more video clips can be used for learning a network model.

\subsection{Feature extraction using layer-dependable TGN}
\textbf{Feature extraction:} The features of each sample clip are extracted by 3D ConvNet with shared parameters. We use three backbones as feature extractors, C3D~\cite{Tran2015}, R3D, and R(2+1)D~\cite{Tran2018} to learn the spatio-temporal features effectively.

Our approach is to stack nine convolution layers in C3D and use the output of the Conv5b layer as a feature. The size of the 3D kernel is 3$\times$3$\times$3. Both R3D and R(2+1)D consist of five blocks of convolution layers. The convolution block of R3D comprises two convolution layers with batch normalization and ReLU activation, respectively. The convolution block of R(2+1)D is similar to that of R3D except for the number of convolution layers. The last layer of the two models uses global adaptive pooling to extract spatio-temporal features.

\begin{figure}[t]
\centering
\includegraphics[width=8.4 cm]{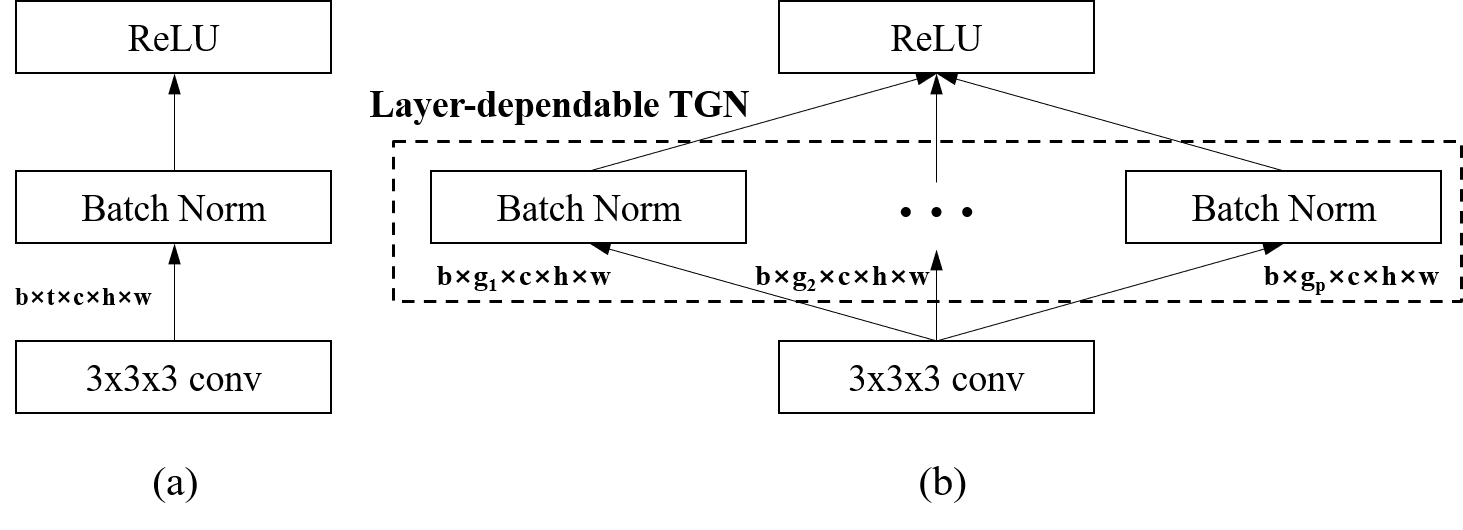}
\caption{\textbf{Network comparison}. (a) Original batch normalization in video processing, and (b) proposed layer-dependable temporal groups normalization.}
\label{fig:03}
\end{figure}

\setlength{\tabcolsep}{4pt}
\begin{table}[t]
\small
\begin{center}
\caption{C3D consists of five layers and the temporal group feature size is $g=2$. After Conv1, Conv2, and Conv 3, each max pooling layer reduces the temporal feature size by half. $t$ is a temporal feature size and $p$ is a number of groups. }
\label{table:01}
\begin{tabular}{c|c|c|c|c|c}
\hline
Layer & Conv1 & Conv2 &Conv3 &Conv4&Conv5\\
\hline
\hline
$t$ & 16& 8& 4& 2& 2\\
\hline
$p$ &8 & 4 & 2 & 1 & 1\\
\hline
\end{tabular}
\end{center}
\vspace{-5mm}
\end{table}
\setlength{\tabcolsep}{1.4pt}

\setlength{\tabcolsep}{4pt}
\begin{table}[t]
\small
\begin{center}
\caption{Performance comparison between simple speed prediction and multiple playback speed order prediction. Both methods use three different speeds.}
\label{table:02}
\begin{tabular}{c|c|c }
\hline
Task	&Speed Pred.	&Playback Speed Order Pred.\\
\hline
\hline
Accuracy&	64.35&	\textbf{69.47}\\
\hline
\end{tabular}
\end{center}
\end{table}
\setlength{\tabcolsep}{1.4pt}

\setlength{\tabcolsep}{4pt}
\begin{table}[t]
\small
\begin{center}
\caption{\textbf{Ablation study of playback directions}. We use C3D with BN as the backbone. The model learns to predict the order of clips played at three different playback speeds. The clips were played back by using fast forward (FF) and rewind (RW) at three different speeds $s\in\{1,3,5\}$ in each playback direction.}
\label{table:03}
\begin{tabular}{c|c|c|c}
\hline
Method	&FF & RW & FF+RW\\
&\{1$\times$, 3$\times$, 5$\times$\}&	\{-1$\times$, -3$\times$, -5$\times$\}&	\{-3$\times$, 1$\times$, 3$\times$\}\\
\hline
\hline
Accuracy	&67.25	&67.75	&\textbf{69.47}\\
\hline
\end{tabular}
\end{center}
\vspace{-5mm}
\end{table}
\setlength{\tabcolsep}{1.4pt}
\setlength{\tabcolsep}{4pt}
\begin{table}[t]
 \small
\begin{center}
\caption{\textbf{Comparison of action recognition accuracy by the playback speeds, $s$}. In order to explore this ablation study, we set $n = 3$ and combination of two direction.}
\label{table:04}
\begin{tabular}{c|c|c|c}
\hline
$s$ &	\{-2$\times$, 1$\times$, 2$\times$\}	&\{-3$\times$, 1$\times$, 3$\times$\}	&\{-4$\times$, 1$\times$, 4$\times$\}\\
\hline
\hline
Accuracy&	67.50&	\textbf{69.47}&	68.81\\
\hline
\end{tabular}
\end{center}
\end{table}
\setlength{\tabcolsep}{1.4pt}

\textbf{Layer-dependable TGN:} In video processing, the dependence between frames is not the same at all times. For example, the frame $f_i$ at a specific time $i$ has a higher correlation with $f_{(i+1)}$ than $f_{(i+k)}$ when $k\gg1$. In other words, as time passes, the correlation of the preceding frame with the current frame is reduced and this characteristic is clearer when the video playback speed is fast. Note that our method uses fast playback as a task of self-supervised learning. 

As shown in Fig~\ref{fig:03} (a), the original batch normalization now obtains the mean and variance from each batch without considering frame changes over time thereby ignoring inherent spatio-temporal characteristics of the video. However, in this paper, we propose a layer-dependable temporal groups normalization method that normalizes individual groups divided along with temporal features, as shown in Fig~\ref{fig:03} (b). Note that the number of temporal groups, $p$, depends on the depth of the corresponding layers, as shown in Table~\ref{table:01}. Because we use the fixed temporal group feature size, $g$, and the temporal feature size, $t$, is expected to change according to the depth of the layers. In detail, The number of the temporal groups is decreased according to the increasing depth of the layers. From this scheme, we consider the many groups of features at the lower layers and the small groups of the feature at the higher layers when learning the 3D ConvNet. The features of the lower layers change with time while the features of the higher layer are processed for the target loss of action recognition. 
%
Its formulations for each channel indexed by $\alpha$ are as follows:
\begin{align}
    \mu _{i} ^{(\alpha )}&=\frac{1} {bg_ihw} \sum\limits_{j \in \{B, G_i, H, W\}}X_j,
    \label{eq:05}\\
    \sigma _{i}^{(\alpha )}&=\sqrt{\frac{1} {bg_ihw}\sum\limits_{j\in\{B,G_i,H,W\}}(x_j - \mu _i)},
    \label{eq:06}
\end{align}
where $X \in \Re^{b\times g_i\times c\times h\times w}$ is the feature computed by a layer which is a 5D vector indexing the features in ($B$, $G_i$, $C$, $H$, $W$), and $b$, $g_i$, $c$, $h$, $w$ denote the feature size of the mini-batch, $i^{th}$ temporal group, channel, height, and width, respectively. Furthermore, $T=\{G_1,...,G_p\}$ when $T$ is the temporal feature, $G_i$ is $i^{th}$ temporal group feature, and $p$ is the total number of temporal groups.
Our proposed TGN computes $\mu$ and $\sigma$ along the ($B$, $G_i$, $H$, $W$) axes for each channel. The general formulation of TGN feature normalization is as follows: 
\begin{equation}
    y_i^{(\alpha)}=\gamma_i^{(\alpha)} \hat{x_i}+\beta_j^{(\alpha)}, 
\end{equation}
where $\hat{X_i}^{(\alpha)}=\frac{(X_i-\mu_i^{(\alpha)})} {\sigma_i^{(\alpha)}}$. Specifically, pixels in the same group are normalized together by $\mu$ and $\sigma$. The TGN also learns the values of $\gamma$ and $\beta$ for each channel.

\subsection{Playback speed order prediction}
After extracting the features from the 3D ConvNet and the proposed TGN method, they are concatenated pairwise as shown in Fig~\ref{fig:02} (c). We use a multi-layer perceptron to encode the pairwise concatenated features, which is a straightforward architecture for solving the order prediction problem~\cite{HLee2017, DXu2019}. The final order prediction is then formulated using the softmax function with the concatenation of all pairwise features. We follow the protocol in \cite{HLee2017} and in a single optimization step we always apply and predict all combinations for every clip in a mini-batch. Compared with \cite{HLee2017, DXu2019}, the main difference is that we predict the order of multiple playback speeds instead of the order of frames or clips.

\section{Experimental Results and Discussion}

\textbf{Datasets:} We evaluate our methods on two action recognition datasets, UCF-101~\cite{ucf} and HMDB-51~\cite{hmdb} from the viewpoint of classification accuracy and retrieval performance. The UCF-101 dataset consists of 13,320 videos obtained from YouTube. These videos are classified into 101 classes. HMDB-51 is a dataset that expresses 51 human actions at least 101 times per class and consists of the total 6,849 clips. In this paper, we verify our method by using the clips for pre-training to compare the performance of the self-supervised learning methods~\cite{HLee2017, DXu2019, DLou2020}. Specifically, we train a 3D ConvNet using UCF-101 without label information first and fine-tune the model using labeled videos such as UCF-101 and HMDB-51, respectively. The network output is a 512-dimensional vector after the global spatio-temporal pooling layer, to which we append a fully-connected layer with softmax on top of it, as in~\cite{DXu2019} in the action recognition. The appended layer is only randomly initialized and the other layers are initialized from the self-supervised learning task. 

\textbf{Implementation details:} All experiments are conducted using PyTorch~\cite{pytorch}. We employ three well-known backbones, that is, C3D, R3D, and R(2+1)D. 
The input clips are resized and randomly cropped to 112$\times$112.
The clips are cropped in the center during the test procedure. When sampling the frames from a video, the starting position is set randomly and a total of $m=$16 frames are extracted. To ensure that the 3D ConvNet is efficiently trained, which may be problematic in terms of memory consumption, we adopt SGD training on one GPU. The mini-batch size is 8$\times n$, where $n$ denotes the number of clips. The process of training the proposed PSP network continue for 300 epochs, following by another 150 epochs of fine-tuning to train the network to perform action recognition at a learning rate of 0.001. We use the momentum of 0.9, and weight decay of 0.0005. A dropout rate of 0.5 is used before the final fully connected layer.

\setlength{\tabcolsep}{4pt}
\begin{table}[t]
\small
\begin{center}
\caption{\textbf{Performance changes with the number of clips}. We use both forward and backward playback directions. VCOP is Video Clip Order Prediction~\cite{DXu2019}. For example, 2, 3, 4, 5, and 6 clips have $\{-3\times, 3\times\}$, $\{-3\times,1\times,3\times\}$, $\{-3\times,-1\times,1\times,3\times\}$, $\{-5\times,-3\times,1\times,3\times,5\times\}$, and $\{-5\times,-3\times,-1\times,1\times,3\times,5\times\}$, respectively.}
\label{table:05}
\begin{tabular}{c|c|c|c|c|c}
\hline
\# of clips & 2 Clips &3 Clips&4 Clips & 5 Clips & 6 Clips\\
\hline
\hline
Ours&	65.56&	69.47&	69.10&	\textbf{71.70}&	71.40\\
\hline
VCOP&	67.80&	67.93 & 66.77&	61.67&	-\\
\hline
\end{tabular}
\end{center}
\vspace{-5mm}
\end{table}
\setlength{\tabcolsep}{1.4pt}

\begin{table}[t]
\setlength{\tabcolsep}{4pt}
    \small
    \begin{center}
    \caption{\textbf{Performance variation according to the number of elements in each groups}. $m=16$ and 3 clips are used. 
    }
    \label{table:06}
    \begin{tabular}{c|c|c|c|c|c}
    \hline
    Model & \multicolumn{5}{c}{The element size of groups, $g$}\\
    \cline{2-6}
    &16(=BN)	&8	&4	&2	&1\\
    \hline
    \hline
    C3D&	69.47&	70.05&	69.84	&\textbf{70.26}&	70.05\\
    \hline
    R3D&	67.43&	67.64&	65.58	&\textbf{69.10}&	67.06\\
    \hline
    R(2+1)D&	70.74&	70.36 &	71.72	&\textbf{72.92}&72.51\\
    \hline
\end{tabular}
\end{center}
\end{table}
\setlength{\tabcolsep}{1.4pt}

\begin{table}[t]
\setlength{\tabcolsep}{4pt}
\small
\begin{center}
\caption{\textbf{Performance variation according to the number of clips}. The effect of increasing the number of clips on the performance for $g=2$ is determined.}
\label{table:07}
\begin{tabular}{c|c|c|c|c}
\hline
Model	&	3 clips & 4 clips	&	5 clips & 6 clips\\
\hline
\hline
C3D	&70.26& 70.34&	\textbf{71.53}& 69.51\\
\hline
R3D	&69.10& 69.42&	\textbf{69.44} & 67.06\\
\hline
R(2+1)D & 72.92 & 73.01 & \textbf{74.65} & 73.67 \\
\hline
\end{tabular}
\end{center}
\vspace{-5mm}
\setlength{\tabcolsep}{1.4pt}
\end{table}

\setlength{\tabcolsep}{4pt}
\begin{table}[t]
 \small
\begin{center}
\caption{\textbf{Action recognition accuracy of variant 3D ConvNet based methods on HMDB-51 and UCF-101.} The average accuracy is measured over three splits.
}
\label{table:08}
\begin{tabular}{c|c|c|c|c|c|c}
\hline
Dataset &	\multicolumn{3}{c|}{HMDB-51} &\multicolumn{3}{c}{UCF-101}	\\
\hline
Model	& C3D&	R3D&	R(2+1)D&	C3D&	R3D&	R(2+1)D\\
\hline
\hline
 Random & 24.7&	23.4&	22.0&	61.8&	54.5&	55.8\\
\hline
VCOP&	28.4&	29.5&	30.9&	65.6&	64.9	&72.4\\
\hline
VCP&	32.5&	31.5&	32.2&	68.5&	66.0&	66.3\\
\hline
\hline
Ours&	\textbf{34.31}&	\textbf{33.68}&	\textbf{36.82}&	\textbf{70.44}&	\textbf{68.98}&	\textbf{74.82}\\
\hline
\end{tabular}
\end{center}
\end{table}
\setlength{\tabcolsep}{1.4pt}

\setlength{\tabcolsep}{4pt}
\begin{table*}[t]
\centering
\begin{center}
\caption{\textbf{Comparison with the published works on self-supervised learning using visual information}. We indicate random cropping, scaling, random horizontal flip, and color jittering abbreviated as RC, S, HF, and CJ in the augmentation column, respectively. The number after the R3D is the number of the layers. The highest accuracy in each backbone architecture is indicated in bold.}
\label{table:SOTA}
\begin{tabular}{c|c|c|c|c|c|c|c|c|c|c}
\hline
\multirow{2}{*}{Method}                & \multirow{2}{*}{Backbone}    & \multicolumn{4}{c|}{Augmentation}  & \multirow{2}{*}{Input size} & \multirow{2}{*}{Frames}                & \multirow{2}{*}{Train DB}    & \multirow{2}{*}{UCF-101}           & \multirow{2}{*}{HMDB-51} \\
\cline{3-6}
                                       &                              & RC     & S      & HF     & CJ      &                             &
                                       &                              &                                    &                           \\
\hline
\hline
SpeedNet\cite{speednet2020}            & \multirow{3}{*}{S3D}         & \cmark &        &        &\cmark   & 224                         & 64     
                                       & Kinetics                     & 81.1                               & 48.8    \\
\cline{1-1} \cline{3-11}
PP\cite{ppnet2020}                     &                              & \cmark & \cmark & \cmark &\cmark   & 224                         & 64
                                       & UCF-101                      & 87.1                               & \textbf{52.6}    \\
\cline{1-1} \cline{3-11}
CoCLR\cite{coclr2020}                  &                              & \cmark & \cmark & \cmark &\cmark   & 128                         & 32
                                       & UCF-101                      & 81.4                               & 52.1    \\
\cline{1-1} \cline{3-11}
Ours                                   &                              & \cmark & \cmark & \cmark &\cmark   & 128                         & 32
                                       & Kinetics                     & \textbf{88.7}                      & \textbf{54.9}    \\
\hline
\hline
VCOP\cite{DXu2019}                     &\multirow{3}{*}{C3D}          & \cmark &        &        &         & 112                         & 16
                                       & UCF-101                      & 65.6                               & 28.4    \\
\cline{1-1} \cline{3-11}
VCP\cite{DLou2020}                     &                              & \cmark &        &        &         & 112                         & 16
                                       & UCF-101                      & 68.5                               & 32.5    \\
\cline{1-1} \cline{3-11}
\textbf{Ours}                                   &                              & \cmark &        &        &         & 112                         & 16
                                       & UCF-101                      & \textbf{70.4}                      & \textbf{34.3}    \\
\Xhline{3\arrayrulewidth}
VCOP\cite{DXu2019}                     & \multirow{3}{*}{R3D-10}      & \cmark &        &        &         & 112                         & 16     
                                       & UCF-101                      & 64.9                               & 29.5    \\
\cline{1-1} \cline{3-11}
VCP\cite{DLou2020}                     &                              & \cmark &        &        &         & 112                         & 16
                                       & UCF-101                      & 66.0                               & 31.5    \\
\cline{1-1} \cline{3-11}
\textbf{Ours}                                   &                              & \cmark &        &        &         & 112                         & 16     
                                       & UCF-101                      & \textbf{69.0}                      & \textbf{33.7}    \\
\Xhline{3\arrayrulewidth}
ST-Puzzle\cite{DKim2019}               & \multirow{5}{*}{R3D-18}      & \cmark & \cmark & \cmark &         & 224                         & 16     
                                       & Kinetics                     & 65.8                               & 33.7    \\
\cline{1-1} \cline{3-11}
\multirow{2}{*}{DPC\cite{DPC2019}}     &                              & \cmark & \cmark & \cmark &         & 128                         & 25     
                                       & UCF-101                      & 60.6                               & -       \\
\cline{3-11}
                                       &                              & \cmark & \cmark & \cmark &         & 128                         & 25     
                                       & Kinetics                     & 68.2                               & 34.5    \\
\cline{1-1} \cline{3-11}
TT\cite{temporaltransforms2020}        &                              & \cmark & \cmark & \cmark &\cmark   & 128                         & 16     
                                       & Kinetics                     & 79.3                               & \textbf{49.8}    \\
\cline{1-1} \cline{3-11}
\textbf{Ours}                                   &                              & \cmark & \cmark & \cmark &\cmark   & 128                         & 16     
                                       & Kinetics                     & \textbf{82.8}                      & 48.8    \\
\Xhline{3\arrayrulewidth}
VCOP\cite{DXu2019}                     & \multirow{6}{*}{R(2+1)D}     & \cmark &        &        &         & 112                         & 16
                                       & UCF-101                      & 72.4                               & 30.9    \\
\cline{1-1} \cline{3-11}
VCP\cite{DLou2020}                     &                              & \cmark &        &        &         & 112                         & 16     
                                       & UCF-101                      & 66.3                               & 32.2    \\
\cline{1-1} \cline{3-11}
\textbf{Ours}                                   &                              & \cmark &        &        &         & 112                         & 16     
                                       & UCF-101                      & \textbf{74.8}                      & \textbf{36.8}    \\
\cline{1-1} \cline{3-11}
PP\cite{PacePrediction}                &                              & \cmark & \cmark & \cmark &\cmark   & 112                         & 25     
                                       & Kinetics                     & 77.1                               & 36.6    \\
\cline{1-1} \cline{3-11}
TT\cite{temporaltransforms2020}        &                              & \cmark & \cmark & \cmark &\cmark   & 128                         & 16   
                                       & UCF-101                      & 81.6                               & 46.4    \\
\cline{1-1} \cline{3-11}
\textbf{Ours}                                   &                              & \cmark & \cmark & \cmark &\cmark   & 128                         & 16     
                                       & Kinetics                     & \textbf{83.0}                      & \textbf{50.2}    \\

\Xhline{3\arrayrulewidth}
Supervised                             & R3D-18                       & \cmark & \cmark & \cmark &         & 112                         & 16    
                                       & Kinetics                     & 83.5                               & 53.6   \\
\hline
\end{tabular}
\end{center}
\vspace{-5mm}
\end{table*}
\setlength{\tabcolsep}{1.4pt}

\subsection{Ablation Test}
This section describes the validation of our method based on the C3D CNN using split 1 of UCF-101 for classification.

\textbf{Playback speed prediction network:} We first investigate whether the prediction of the order of the variable playback speed clips differed from simple speed prediction. The simple speed prediction is performed using softmax, and a method of which the backbone architecture is the same as that of the proposed method is used. The total number of different speeds is three and the target speed ranging from $-3\times$ to $3\times$ is set randomly each iteration. Our method uses three different playback speeds such as $\{-3\times, 1\times, 3\times\}$. Table~\ref{table:02} compares the performance of two tasks. The accuracy of the proposed method is 5\% higher as a result of learning more spatio-temporal information using several clips per iteration. 
Note that we cannot accurately predict the playback speed of the video by just watching the video; however, if the playback speeds of videos were to differ, we would easily be able to determine which video is faster or slower. In this respect, order prediction is a more appropriate task than speed prediction for self-supervised learning. 

Next, we aim to explore the interrelation between playback direction and performance. The action recognition accuracy for the fast forward (67.25\%), rewind (67.75\%), and the combination of both (69.47\%) of these playback methods in both directions are listed in Table~\ref{table:03}. These results indicate that learning only single direction provide a similar accuracy, but the combined result is superior. Therefore, we can infer that the different playback directions provide a chance to learn different visual dynamics during training. For example, because walking backward involves the use of different muscles compared to walking forward~\cite{BFWalk}, rewinding a video in which a person is walking forward would be different from playing a video in which someone is walking backward. In other words, rewinding the video could provide richer visual information by playing the fast forward. Consequently, our proposed method offers a way to learn more information at once by learning in both directions at the same time.

We test the effect of different playback speeds to determine the greatest relative speed difference for the best performance. As is clear from Table~\ref{table:04}, an extremely small difference between playback speeds results in poor accuracy because the temporal dynamics are not significantly different between the 1$\times$ and 2$\times$ playback speeds. Moreover, learning from an excessively large difference in speed also does not benefit. Our method performs optimally when $s=3$.

Finally, we evaluate different methods to understand the effect of the number of clips to determine the effect of variable $n$. The results in Table~\ref{table:05} indicate that the proposed method, the accuracy increased as the number of clips increased, but decreased again when six clips were used. This is largely because the number of possible orders in which the clips are arranged increased significantly; for example, for five and six clips have $5!=120$ and $6!=720$, respectively, in which case the complexity of the task is too large to train the model successfully. Compared with the VCOP method~\cite{DXu2019}, The performance of the proposed method is superior ranging from 2 to 5 clips as a tuple set. Specifically, the best accuracy, 67.93\%, of VCOP at three clips is approximately 3.77\% lower than the best performance overall, 71.70\%, achieved by the proposed method for five clips. On the basis of this result, we conclude that the proposed method is more effective than the previous methods for different numbers of clips.


\textbf{Layer-dependable TGN:} We evaluate the performance improvement by the proposed TGN using backbones. We determine the optimal number of elements in the groups, and the results are presented in Table~\ref{table:06}. All backbones achieve the best accuracy with $g=2$ to normalize the output of each layer. Most of the results in Table~\ref{table:06} are higher (i.e., more accurate) than the BN baseline except for $g=1$. Note that, at R(2+1)D, the proposed TGN improved the accuracy by 2.18\% compared with the BN baseline.
However, in case of C3D, performance improvement is marginal compared to R3D and R(2+1)D. C3D is difficult to train temporal representation because appearance and dynamics are jointly intertwined~\cite{Tran2018}.
Therefore, TGN is more effective for R3D and R(2+1)D that can extract appearance  and dynamics separately because TGN emphasizes temporal representation but cannot extract features.

Table~\ref{table:07} lists the effect of the number of clips on the action recognition performance, and the accuracy of the task using five clips is superior compared with that using three clips for backbones. Notably, the performance is saturated when the number of clips is over 5. The reason is that as the number of clips increases, the burden on the order prediction network increases.
In conclusion, the proposed TGN if more effective for larger variations in the dynamics and appearance of the sampled clips, which is the consequence of the task responsible for the proposed PSP network.

\subsection{Action Recognition}

In this section, we shows the effectiveness of our method in action recognition. We conduct experiments on UCF-101 and HMDB-51 and results are measured over three split of each dataset. In Table~\ref{table:08}, we compare the proposed method with the latest self-supervised learning methods such as VCOP \cite{DXu2019} and VCP~\cite{DLou2020} for both UCF-101 and HMDB-51. To validate the generality of the proposed method, we evaluate three backbone networks: C3D, R3D, and R(2+1)D in these experiments. Compared with random initialization (e.g., training from scratch) for 3D ConvNets, all self-supervision methods including ours exhibit superior performance. However, our method always outperforms the two other methods such as VCOP and VCP regardless of the 3D ConvNet models.

From table~\ref{table:SOTA}, our method shows a comparable results compared to the state-of-the-art (SOTA) self-supervised methods such as VCOP, VCP, Dense Predictive Coding (DPC)~\cite{DPC2019}, SpeedNet~\cite{speednet2020}, Temporal Transformation (TT)~\cite{temporaltransforms2020}, Pace Prediction (PP)~\cite{PacePrediction}, and CoCLR~\cite{coclr2020}.
Since each method has different backbone architecture, hyper-parameter setting, and augmentation method, we compare the performance of each backbone and augmentation setting to ensure fair comparison.
Our method achieves the best results on almost backbones and augmentation settings over two datasets.
The use of random crop only as an augmentation method is difficult to show good performance because of the high possibility of training shortcut, but our method shows about 2\% to 5\% improvement in all backbone compared to VCOP or VCP.
In addition, our method shows higher accuracy than other methods using the same augmentation even when training shortcut is removed using color jittering or horizontal flip.
To be noticed, our performance of R3D-18 as backbone is close to the supervised model using the pretrained on Kinetics.
In addition, we make the best accuracy using S3D on UCF-101 and HMDB-51.


\begin{figure*}[t]
\centering
\includegraphics[height=9cm]{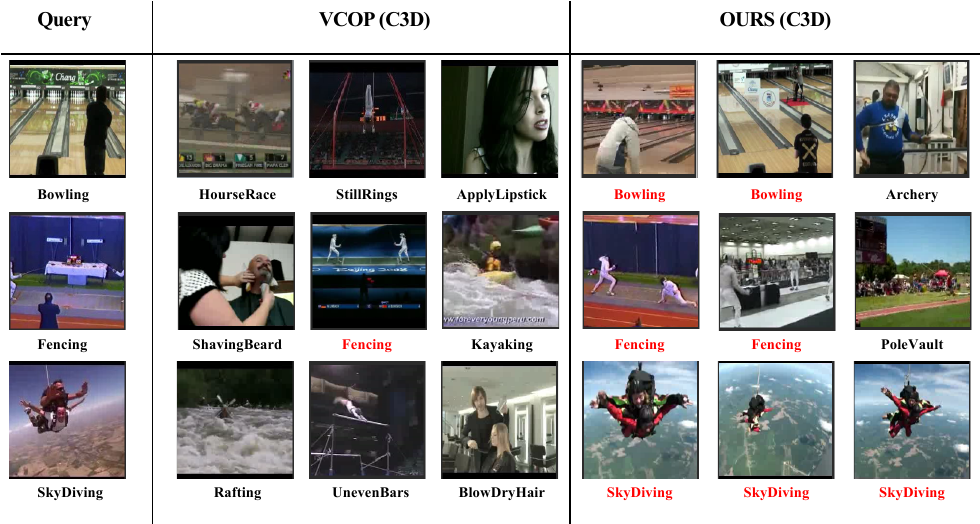}
\caption{\textbf{Samples of video clip retrieval results}. The labels highlighted in red indicate that this video clip is in the same category as the test video clip.}
\label{fig:04}
\end{figure*}

\begin{figure}
\centering
\includegraphics[height=9.5cm]{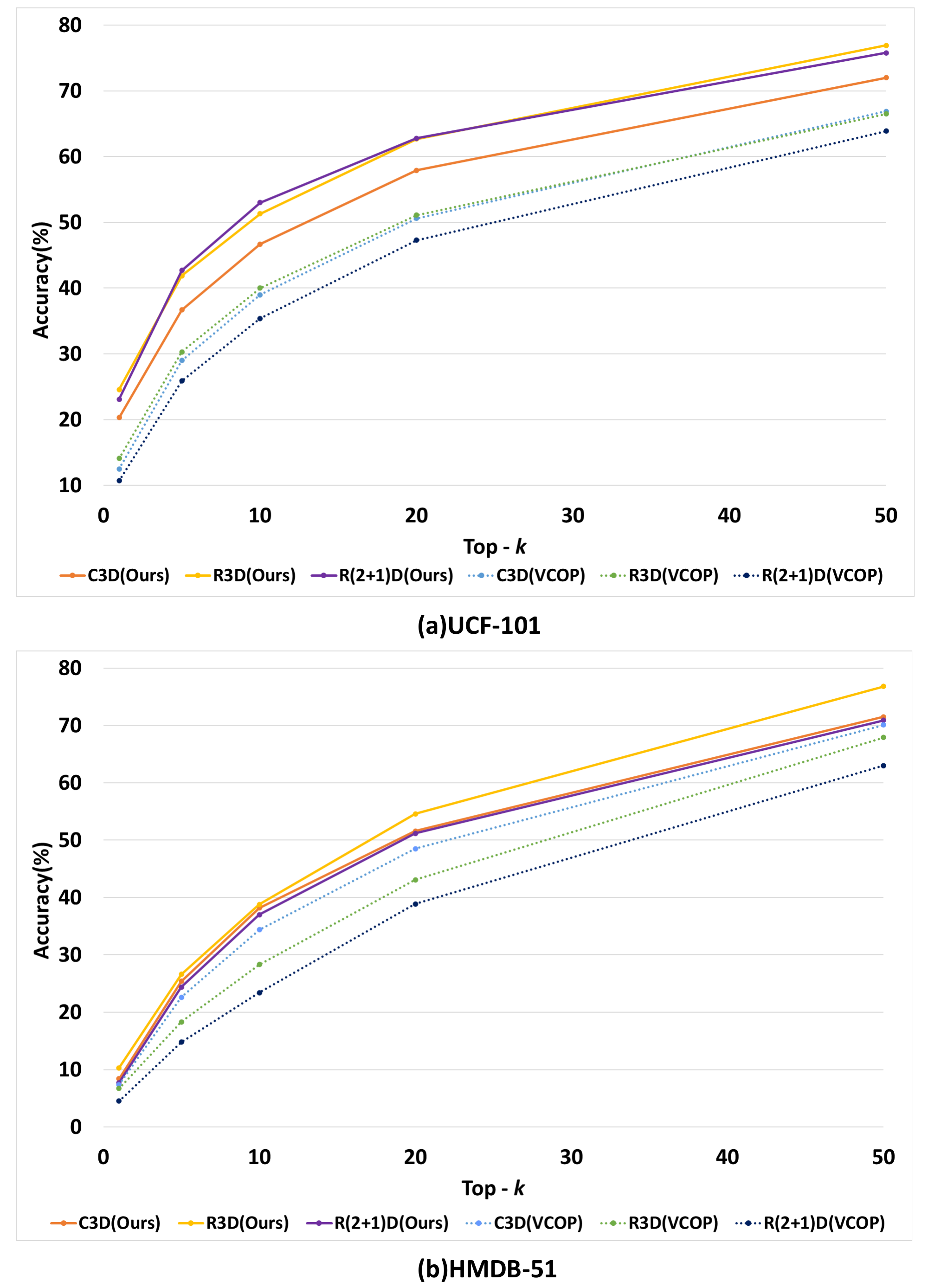}
\caption{\textbf{Results of video retrieval}. Our performances are compared with those of VCOP for C3D, R3D, and R(2+1)D networks.}
\label{fig:05}
\end{figure}

\subsection{Video Clip Retrieval}

\setlength{\tabcolsep}{4pt}
\begin{table}[t]
\small
\centering
\caption{\textbf{Video clip retrieval results on UCF-101 and HMDB-51}. The backbone architectures of VCOP, VCP, and ours are 3D ConvNet based on the self-supervised learning method and we chose the best accurate network among C3D, R3D, and R(2+1)D. 
}
\label{table:10}



\begin{tabular}{c|c|c|c|c|c|c}

\hline
\multirow{2}{*}{DB}      & \multirow{2}{*}{Method} & \multicolumn{5}{c}{Top-K}                                                    \\ \cline{3-7} 
                         &                         & 1             & 5             & 10            & 20            & 50            \\ \hline
\hline
\multirow{5}{*}{UCF-101} & VCOP                    & 14.1          & 30.3          & 40.4          & 51.1          & 66.5          \\ \cline{2-7} 
                         & VCP                     & 19.9          & 33.7          & 42.0          & 50.5          & 64.4          \\ \cline{2-7} 
                         & SpeedNet                & 13.0          & 28.1          & 37.5          & 49.5          & 65.0          \\ \cline{2-7} 
                         
                         & \textbf{Ours}           & \textbf{24.6} & \textbf{41.9} & \textbf{51.3} & \textbf{62.7} & \textbf{67.9} \\ \hline
\multirow{3}{*}{HMDB-51} & VCOP                    & 7.6           & 22.9          & 34.4          & 48.8          & 68.9          \\ \cline{2-7} 
                         & VCP                     & 7.8           & 23.8          & 35.3          & 49.3          & 71.6          \\ \cline{2-7} 
                         & \textbf{Ours}           & \textbf{10.3} & \textbf{26.6} & \textbf{38.8} & \textbf{54.6} & \textbf{76.8} \\ \hline
\end{tabular}
\end{table}
\setlength{\tabcolsep}{1.4pt}\textbf{}

The performance of our method is confirmed by searching for nearest-neighbor video retrieval. The overall process of video retrieval used in the experiment followed that of \cite{Buchler18}\cite{DXu2019} and was evaluated with the split 1 of UCF-101 and HMDB-51 as in the previous papers. The first step of the entire experimental process entailed loading the weight of the trained model by using the training protocol presented in this paper. At this time, the feature was extracted by using 3D max pooling instead of the pre-existing global spatio-temporal pooling after passing through the final convolutional layer of the weight-loaded model. To measure the retrieval performance, we calculate the cosine distance between the testing and training video sets. The shortest nearest-neighbor video clip of $k$ is found among the calculated cosine distances. If the correct answer is included among the $k$ nearest neighbor video clips, the result is considered to be successful. After the video retrieval is performed for all test video sets, the accuracy is calculated by obtaining the number of correct answers divided by the total. In Fig~\ref{fig:04}, selected results of the video clip retrieval through samples are shown. Compared with the VCOP method, the qualitative results obtained with our method for the direction dynamics of videos such as bowling, fencing, and skydiving are more accurate. 

The quantitative results are shown in Table~\ref{table:10}. The proposed method achieves the highest accuracy among the well-known self-supervised methods~\cite{Noroozi2016}\cite{HLee2017}\cite{Buchler18}\cite{DXu2019}\cite{DLou2020} through most ranks ranging from 1 to 50 at both UCF-101 and HMDB-51 databases. Specifically, the proposed method shows up to 5.7\% and 2.8\% higher accuracy than the state-of-the-art methods at the top-5 rank of UCF-101 and HMDB-51 datasets, respectively. Regardless of the types of the 3D ConvNets, this superiority is also confirmed by the ROC curves in Fig~\ref{fig:05}. We compared the results with VCOP for various 3D ConvNet architectures, the solid line is the PSP results that we propose, and the dotted line is the VCOP results. (a) shows a visualize for the top-k retrieval result in UCF-101 database and (b) shows a visualize for the top-k retrieval result in HMDB-51 database.  From this result, we conclude that the proposed method works better than the well-known methods in the video retrieval task.

\section{Conclusions}
In this paper, We propose a various playback speed prediction network as a salient surrogate of the self-supervised learning and a layer-dependable temporal group normalization for 3D ConvNet. Our method is able to train the natural visual temporal flow of videos as a pre-trained model by ordering the different fast forward playback speeds as well as the rewind speeds and designed the novel temporal group normalization for efficient visual representation learning at action recognition and video retrieval task. In the end, we have verified the superiority of our method by the extensive experiments.

\bibliographystyle{unsrt}
\bibliography{reference}

\begin{IEEEbiography}[{\includegraphics[width=1in,height=1.25in,clip,keepaspectratio]{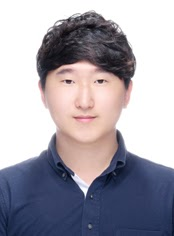}}]{Hyeon Cho} received B.S. degree from the Department of Software and Computer Engineering, Ajou University, Korea, in 2018, and now pursuit Ph.D degree. He is currently studying on improving the performance of the action recognition model. His current research interests include computer vision, pattern recognition and deep learning.
\end{IEEEbiography}

\begin{IEEEbiography}[{\includegraphics[width=1in,height=1.25in,clip,keepaspectratio]{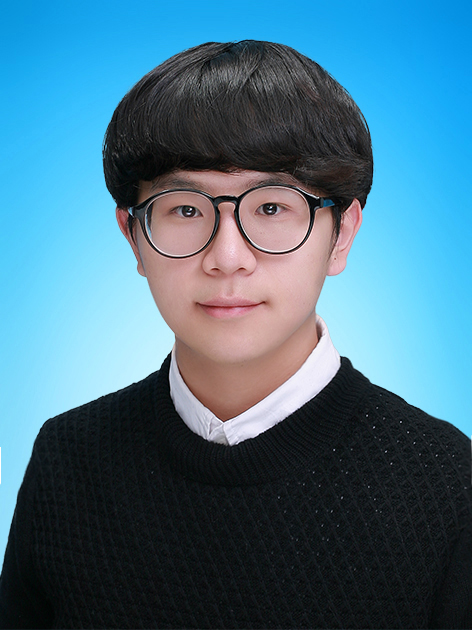}}]{Taehoon Kim}  received B.S. degree from the Department of Software and Computer Engineering, Ajou University, Korea, in 2020, and now pursuit Master's degree. He is currently studying on Video Recognition using Self-Supervision manners. His current research interests include computer vision, pattern recognition and deep learning.
\end{IEEEbiography}

\begin{IEEEbiography}[{\includegraphics[width=1in,height=1.25in,clip,keepaspectratio]{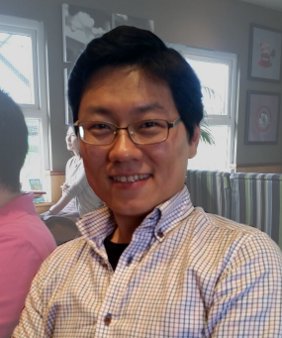}}]{Hyung Jin Chang} received his B.S. and Ph.D. degree from the School of Electrical Engineering and Computer Science, Seoul National University, Seoul, Republic of Korea. He was a post doctoral researcher with the Imperial Computer Vision and Learning Lab and the Personal Robotics Laboratory at the Department of Electrical and Electronic Engineering at Imperial College London. He is currently a lecturer (equivalent to assistant professor) of the School of Computer Science at the University of Birmingham. His current research interests are human understanding through visual data analysis including human/hand pose estimation, eye gaze tracking,  articulated structure learning, human robot interaction, 6D object pose tracking, human action understanding, and user modelling.


\end{IEEEbiography}

\begin{IEEEbiography}[{\includegraphics[width=1in,height=1.25in,clip,keepaspectratio]{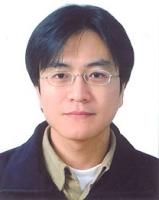}}]{Wonjun Hwang} received both B.S. and M.S. degrees from the Department of Electronics Engineering, Korea University, Korea, in 1999 and 2001, respectively, and Ph.D. degree in the School of Electrical Engineering, Korea Advanced Institute of Science and Technology (KAIST), Korea, in 2016. From 2001 to 2008, he was a research staff member in Samsung Advanced Institute of Technology (SAIT), Korea. In 2004, he contributed to the promotion of Advanced Face Descriptor, Samsung and NEC joint proposal, to MPEG-7 international standardization. In 2006, he proposed the SAIT face recognition method which achieved the best accuracy under the uncontrolled illumination situation at Face Recognition Grand Challenge (FRGC) and Face Recognition Vendor Test (FRVT). In 2006, he developed the real-time face recognition engine for the Samsung cellular phone, SGH-V920. From 2009 to 2011, he was a senior engineer in Samsung Electronics, Korea, where he worked on developing face and gesture recognition methods for Samsung humanoid robot, a.k.a RoboRay. In 2011, he rejoined the SAIT as a research staff member and from 2011 to 2014 he worked for a 3D medical image processing of Samsung surgical robot. From 2014 to 2016, he worked on developing deep learning-based face detection and recognition methods for Samsung Galaxy series. In 2016, he joined the department of Software and Computer Engineering, Ajou University, Korea and is now an associate professor. His research interests are in computer vision, pattern recognition, and deep learning.

\end{IEEEbiography}
\EOD
\end{document}